\documentclass[conference]{IEEEtran}
\IEEEoverridecommandlockouts
\usepackage{cite}
\usepackage{amsmath,amssymb,amsfonts}
\usepackage{algorithmic}
\usepackage{graphicx}
\usepackage{textcomp}
\usepackage{xcolor}
 \usepackage{multirow}
 \usepackage{booktabs}
\def\BibTeX{{\rm B\kern-.05em{\sc i\kern-.025em b}\kern-.08em
    T\kern-.1667em\lower.7ex\hbox{E}\kern-.125emX}}
\begin{document}

\title{AntiDote: Attention-based Dynamic Optimization \\ for Neural Network Runtime Efficiency\\
}

\author{\IEEEauthorblockN{Fuxun Yu$^1$, Chenchen Liu$^2$, Di Wang$^3$, Yanzhi Wang$^4$, Xiang Chen$^1$}
\textit{$^1$George Mason University, Fairfax, VA, USA} \\
\textit{$^2$University of Maryland, Baltimore County, Baltimore, MD, USA} \\
\textit{$^3$Microsoft, Redmond, WA, USA} \\
\textit{$^4$Northeastern University, Boston, MA, USA} \\
\textit{\{fyu2, xchen26\}@gmu.edu, ccliu@umbc.edu, wangdi@microsoft.com, yanz.wang@northeastern.edu}
}

\maketitle

\begin{abstract}
Convolutional Neural Networks (CNNs) achieved great cognitive performance at the expense of considerable computation load.
	To relieve the computation load, many optimization works are developed to reduce the model redundancy by identifying and removing insignificant model components, such as weight sparsity and filter pruning.
	However, these works only evaluate model components’ static significance with internal parameter information, ignoring their dynamic interaction with external inputs.
	With per-input feature activation, the model component significance can dynamically change, and thus the static methods can only achieve sub-optimal results.
	Therefore, we propose a dynamic CNN optimization framework in this work.
	Based on the neural network attention mechanism, we propose a comprehensive dynamic optimization framework including (1) testing-phase channel and column feature map pruning, as well as (2) training-phase optimization by targeted dropout.
	Such a dynamic optimization framework has several benefits: 
		(1) First, it can accurately identify and aggressively remove per-input feature redundancy with considering the model-input interaction; 
		(2) Meanwhile, it can maximally remove the feature map redundancy in various dimensions thanks to the multi-dimension flexibility;
		(3) The training-testing co-optimization favors the dynamic pruning and helps maintain the model accuracy even with very high feature pruning ratio.
	Extensive experiments show that our method could bring 37.4\%$\sim$54.5\% FLOPs reduction with negligible accuracy drop on various of test networks.
\end{abstract}


\section{\bf{Introduction}}
\label{sec:intro}

In the past few years, Convolutional Neural Networks (CNNs) have achieved extrodinary accuracy boost on various cognitive tasks, such as image classification~\cite{img_cls}, object detection~\cite{obj_det}, speech recognition~\cite{speech}, \textit{etc}.
	However, the increasingly larger model structure also introduces tremendous computation load, causing considerable performance issues.
	For example, a CNN model for natural language processing (NLP) --- \textit{BERT\_Large} consists of 345 million parameters. Its huge memory occupancy makes it even impossible to be trained on a single V100 GPU with 16GB memory~\cite{bert}.

	To reduce the CNN computation load, many optimization works have been proposed to reduce the model redundancy by identifying and removing insignificant components.
	Han \textit{et al.} considered the weight magnitude as the significance criteria and introduced weight sparsity regularization for model compression~\cite{sparse}.
	To improve the sparsity irregularity in~\cite{sparse}, structural pruning was also proposed to achieve practical computation acceleration~\cite{yanzhi,wenwei}.
	Li \textit{et al.} defined the $\ell_1$-norm of the convolutional filter weights for significance evaluation and removed small filters for less computation load~\cite{filter_prune, channel_prune}.


\begin{figure*}[!t]
	\centering
	\vspace{-3.5mm}
	\includegraphics[width=6.3in]{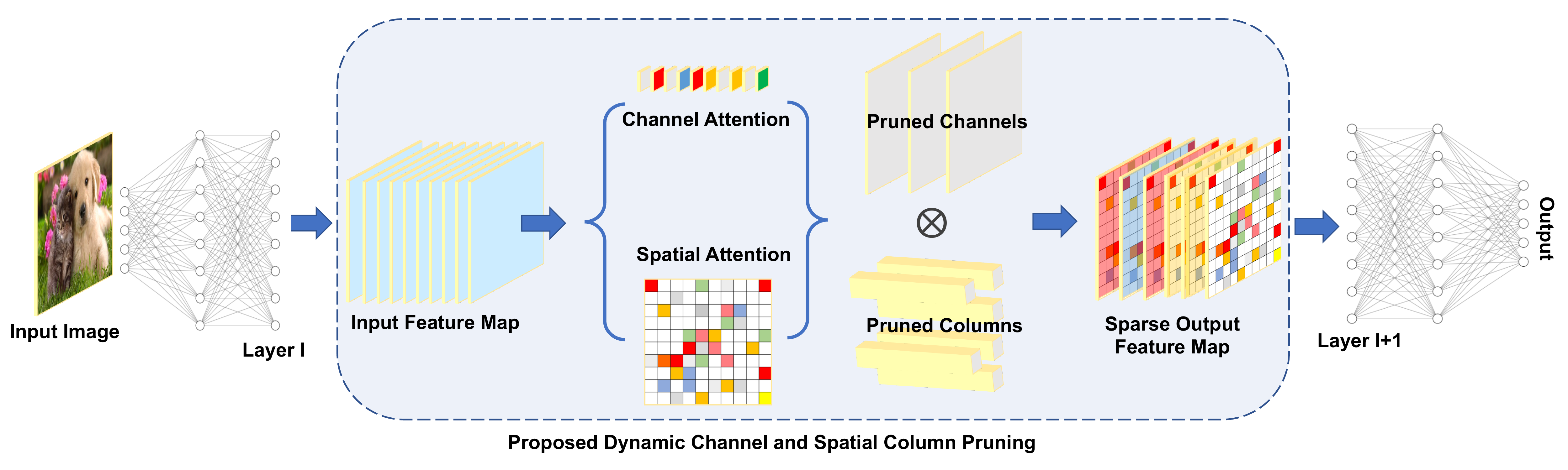}
	\vspace{-3mm}
	\caption{Channel and Spatial Attention-based Dynamic Pruning Framework. Deeper colors denote the channel or spatial column has larger attention coefficients and will be reserved. Other channels and columns will be regarded as redundancy and be removed and the computation related can be thus skipped for efficiency.}
	\label{fig:overview}
	\vspace{-5mm}
\end{figure*}

Although these works demonstrate good performance, all these works only evaluate model components' \textit{static} significance with internal parameter information, ignoring their \textit{dynamic} interaction with external inputs.
	As CNN models are trained for discriminating class-level features, every single input may have certain feature activation variance:
	generic significant neurons may not be activated by particular inputs~\cite{l1_revisit},
	while some insignificant neurons with rare feature preference maybe specifically favored~\cite{fo}.
	Such a variance suggests that the model components' significance should be evaluated with a \textit{dynamic} manner through the practical input testing process.


Recently, many research works have leveraged such a \textit{dynamic} manner for CNN model optimization on various component levels.
	On feature map level, the {ImageNet-2017} winner \textit{SENET} found that different inputs have significant impact on the feature channel significance. And dynamic feature channel pruning can achieve outstanding computation acceleration~\cite{senet}.
	On convolutional filter level, Yu~\textit{et al.} also reveal that, regardless of filter weight, certain filters contain particular class-specific features with significant activation impact.
	Therefore, filters with significant weight can be also aggressively pruned according to specific class inputs~\cite{fuxun}.

Compared to the conventional works based \textit{static} component significance, the \textit{dynamic} optimization has two major advantages:
		(1) Qualitatively, it can identify every component's significance more accurately in the practical testing phase by considering the dynamic feature-input interaction.
		(2) Quantitatively, dynamic optimization enables more aggressive redundancy elimination with per-input component significance evaluation, while the static methods can only identify a general redundancy based on the whole dataset.



However, current dynamic CNN optimization works still have certain limitations:
%
	(1) How to achieve an evaluation metric for dynamic significance evaluation?
%
	(2) How to apply the dynamic optimization in different model dimensions, especially the non-structured ones?
%
	(3) How to develop particular dynamic optimization schemes for both testing and training phases to achieve comprehensive performance improvement?
%

To solve these problems, we propose \textit{AntiDote}: an attention-based dynamic CNN optimization framework.
	Specifically, we have the following contribution:
\begin{itemize}
	\item We establish a dynamic feature significance criteria based on the attention mechanism. Focusing on the feature map level, such a criteria could effectively demonstrate per-input feature activation and corresponding model redundancy in computation.
	\item We propose a comprehensive dynamic feature pruning method with two dedicated schemes: channel-wise pruning and spatial column pruning, which can flexibly prune redundant features in different dimensions;
	\item We propose a corresponding attention based training algorithm coupled with targeted dropout, which greatly enhances the model testing-phase accuracy against dynamic pruning hurts.

\end{itemize}

Fig.~\ref{fig:overview} gives a systematical overview of our dynamic optimization method.
	Between any two consecutive convolutional layers, we utilize the attention-based mechanism to evaluate the importance of different components of feature map and then generate the feature pruning mask.
	The mask will be applied back onto the feature map and to remove the channel-wise and spatial-wise redundancy structurally.
	By doing so, we could remove the non-essential computation in the next layer for runtime efficiency optimization.  


Combining attention-based model feature pruning and the comprehensive testing/training optimization methods, the proposed \textit{AntiDote} framework achieves expected computation load reduction.
	Extensive experiments show that our method can bring 37.4\%$\sim$54.5\% FLOPs reduction for different CNN models (VGG and ResNet) and different datasets (CIFAR and ImageNet) with negligible or no accuracy drop, which consistently outperforms state-of-the-art optimization methods.

\section{\bf{Preliminary}}

\subsection{{Dynamic Neural Network Optimization}}


As we mentioned before, static pruning neglects one key factor that the model parameter importance can be different w.r.t different inputs.
	In other words, for every batch of inputs during model inference, there may still remain some non-important parameters or computations due to the limitation of static pruning.
	This provide a new dimension to conduct model computation reduction by dynamic pruning.

Following this lead, runtime neural pruning~\cite{rl} used a reinforcement learning-based method to conduct dynamic channel pruning conditioned on different inputs.
	Another recent work~\cite{iclr} proposed to add a channel-wise sparsity regularization term during training to enable the dynamic channel pruning during inference.
	All these dynamic pruning works focus on reducing the channel of feature maps but again neglect another dimension, which is the spatial dimension.
	Our method consider both channel-wise and spatial-wise feature map redundancy, and thus could achieve better performance.

\subsection{{Neural Network Attention Mechanism}}

Neural network attention mechanism is firstly introduced in the language processing models~\cite{attention} and is soon exploited to help achieve the state-of-the-art performance in many vision tasks, like image classification and object detection.
For example, \textit{SENET}~\cite{senet} proposed to use channel attention to reweigh different channels of feature maps, which won the 1st place in ImageNet-2017 image classification competition.
	In \textit{SENET}, each channel will be predicted with an attention coefficient (between 0 and 1), which will be used to maintain or shrink the values of that channel.
	Based on similar mechanism, the spatial attention mechanism~\cite{spatial_attention} is then introduced to predict and reweigh each spatial column of the feature maps.

From the perspective of models, the attention mechanism uses the predicted coefficients to help the model focus on the important features (e.g. important channels or key spatial locations are multiplied with coefficients close to 1), and therefore are called attention mechanism.
	By contrast, in our redundancy removal context, such attention coefficients could also tell us where the non-important features are (e.g. whose coefficients are close to 0).
	Therefore, we propose the attention-based dynamic feature pruning method.


\section{\bf{Attention-based Testing-Phase Optimization: \\ Dynamic Channel and Spatial Feature Map Pruning}}

In this section, we introduce our testing-phase optimization by attention-based dynamic feature pruning, which consists of two schemes: channel pruning and spatial column pruning.

\subsection{Channel and Spatial Attention Coefficient Formulation}

We first give the formal definition of channel and spatial attention.
	Given an intermediate feature map $F \in R^{~C*H*W}$ (here, $C, H, W$ is the channel depth, height, width of the feature map), the channel attention coefficient can be obtained by calculating the average of feature map entries in the spatial dimension $H*W$:
	\begin{equation}
	\small
	\vspace{-1mm}
		\begin{split}
			A_{channel}(F, c) = \frac {1} {H*W} \sum_{i}^{H}\sum_{j}^{W}{F_c(i, j)},
		\end{split}
	\vspace{-1mm}
		\label{eq:channel_attention}
	\end{equation}
	\normalsize
	where $A_{channel}(F, c)$ is channel attention for the selected channel $c$. 
	For a given feature map $F \in R^{~C*H*W}$, the channel attention $A_{channel}(F)$ is a $C-$dimensional vector, each entry of which corresponds to one channel of the feature map $F$.	
	In a neural network, the channel attention is usually calculated by adding a global average pooling layer behind the convolution and ReLU layer. 

Similarly, the spatial attention coefficient calculates feature map's mean in the dimension of depth $C$:
	\begin{equation}
	\small
		\begin{split}
			A_{spatial}(F, h, w) = \frac {1} {C} \sum_{i}^{C}{F_{h, w}(i)}.
		\end{split}
		\label{eq:spatial_attention}
	\end{equation}
	\normalsize
As such, a feature map $F \in R^{~C*H*W}$ would have a spatial attention matrix $A_{spatial}(F)$ of $H*W$ size (so-called attention heat map), each entry of which corresponds to a spatial location of the feature map.

In previous works~\cite{senet, attention}, such attention coefficients are usually passed through a sigmoid layer to be normalized into a $(0,1)$ range, and then be multiplied back to the feature map. 
	It guides the model to dynamically put better attentions, but it can hardly remove feature components for neural network acceleration. 
	In this work, we propose to overcome the above challenge by using a further-binarized mask to improve the attention mechanism in pruning context.
	Based on this, two dynamic feature map pruning schemes are developed: channel-wise pruning and spatial-wise column pruning.

\subsection{Dynamic Feature Map Pruning}


\subsubsection{Dynamic channel pruning}

During neural network forward phase, a common convolutional layer will generate a 3-dimensional feature map $F \in R^{~C*H*W}$ (the first batch dimension is omitted for simplicity).
	Based on Eq.~\ref{eq:channel_attention}, channel-wise attention vector $A_{channel}(F)$ is obtained to evaluate the importance of each channel.
	With these coefficients, a binary mask $M_{channel}(F)$ will be generated, which is a $C$-length binary vector to determine the redundancy removal. 
The channel-wise mask can be generated by setting the mask of top-k attention entries to be \textit{True} or \textit{False}, as is described by Eq.~\ref{eq:mask}.
The value of $k$ depends on the overall feature map depth and the reserved percentage hyperparamter $p$, which is selected based on our later layer sensitivity analysis. 
\begin{equation}
\small
	\begin{split}
		\medmuskip=-2mu
		M(F, c)= \left\{
		\begin{aligned}
			 True , &~~~\text{If $c \in topk(A_{channel})$}, \\
			 		&~~~k=int(p~*~C); \\
			 False, &~~~\text{Otherwise}.
		\end{aligned}
		\right.
	\label{eq:mask}
	\end{split}
\end{equation}
\normalsize
Here, $topk$ returns to the indexes of all top-k entries in the channel attention vector. 
The channel mask \textit{True} indicates the current feature map channel will be reserved, while the other channels whose masks are set to \textit{False} will be masked out and not participate in the next layer's convolution computation.

Different from static channel pruning that the selected channels are permanently pruned, in our case the pruned channels are dynamically pruned based on the attention w.r.t. the current input. 
More specifically, for other inputs which use this channel, it can be fully recovered by the input dependent new binary mask.
	As such, the proposed method can achieve per-input redundancy removal and thus achieve larger pruning ratio than traditional static methods, as is shown in Sec.~\ref{sec:exp}.

\subsubsection{Dynamic spatial column pruning} 

Similar to the channel pruning, dynamic spatial column pruning removes feature map columns at different spatial locations according to the spatial attention coefficients.
	Based on the spatial attention coefficients, a mask $M_{spatial}(F)$ will also be generated.
	For spatial column pruning, the mask $M_{spatial}(F)$ is a $H*W$ matrix, each entry of which corresponds to reserving or removing a column of the feature map:
	\begin{equation}
	\small
	\begin{split}
		\medmuskip=-2mu
		M(F, h, w)= \left\{
		\begin{aligned}
			 True , &~~~\text{$(h,w) \in topk(A_{spatial})$}, \\
			 		&~~~k=int(p~*~H~*~W); \\
			 False, &~~~\text{Otherwise}.
		\end{aligned}
		\right.
	\label{eq:smask}
	\end{split}
\end{equation}
\normalsize
	$topk$ here returns to all the 2-dimension indexes with the top attention coefficients.
	The same as channel pruning, the spatial mask will be applied onto the feature map and the unimportant columns will be removed from the feature map. 

Note that both of the dynamic channel and spatial column pruning target at pruning redundant feature maps, which is different from previous filter weights pruning: the removed feature map components skip the further convolution operation and hence reduce the computation load in the next layer. 

\subsection{Attention-based vs. Random Pruning Results Analysis}

\begin{figure}[!tb]
	\centering
	\includegraphics[width=3.4in]{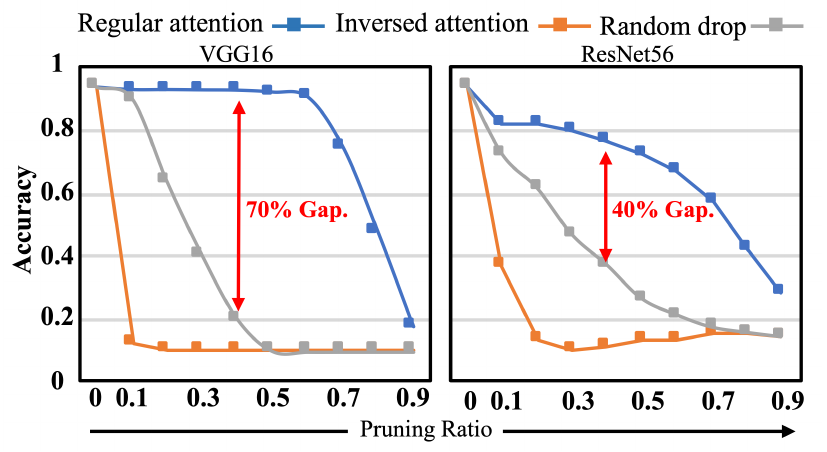}
	\vspace{-3.0mm}
	\caption{Attention-based and Random Pruning Accuracy Drop Comparison.}
	\label{fig:attention_test}
	\vspace{-4mm}
\end{figure}

Attention mechanism has been proved to be effective in improving model performance. However in terms of channel and column pruning, there is no guarantee that it can still provide us good performance. 
	Therefore, in this section, we conduct comparison experiments to evaluate the effectiveness of attention mechanism in dynamic pruning: (1) attention-based pruning, (2) random pruning, and (3) the inversed attention pruning.
	Here the inversed attention pruning means that we firstly prune the channels with the largest attention coefficients, i.e. the opposite priority with attention-based pruning.
	For simplicity, we choose the last block of VGG16 and ResNet56 as the pruning targets. Then dynamic channel pruning based on three criteria is conducted with different pruning ratios to observe the accuracy drop.
	The experimental results are shown in Fig.~\ref{fig:attention_test}.

Clearly, we can find that attention-based pruning outperforms the random pruning by significant large margins. 
	Take the pruning ratio $0.4$ as an example: the accuracy gap between attention-based and random pruning can achieve 70\% and 40\% on VGG and ResNet, which clearly demonstrate the effectiveness of attention coefficients for channel pruning.
	Meanwhile, when we conduct the inversed-attention pruning, the accuracy drops very quickly: 
	For VGG16, we can observe that dropping only 10\% channel will lead to nearly 80\% accuracy drop. 
	This further proves that these top-attention channels are the most essential components, without which the neural network's classification performance can hardly be maintained.
	Similar conclusions could be drawn for dynamic spatial column pruning, which will be demonstrated in the later experiment section.

\section{\bf{Attention-based Training-Phase Optimization: \\ Dynamic Training with Targeted Dropout}}

To relieve the model's dependency on less important feature components, we concurrently propose a optimized training algorithm \textit{TTD: Training with Targeted Dropout}.
	With the new training algorithm, model's accuracy drop resilience against dynamic pruning hurt will be greatly improved, thus providing better dynamic pruning performance. 

\subsection{Training with Targeted Dropout}

The design motivation of \textit{TTD} is to relieve the target model's prediction dependency on less important feature components, so that pruning these components will induce less damage to the model performance.
	Therefore, we propose to integrate dropout mechanism into the model training process. 
	By doing dropout, the model inference process can gradually become robust to the dynamic feature map dropping. 
Meanwhile, to mimic the similar dropping effects as attention-based pruning, the dropout must also target at dropping attention-based less-important feature components.
	Therefore, it's called targeted dropout.
	This is also the main difference between our dropout mechanism from random dropout: the latter one is usually used for totally different purposes, e.g. to avoid over-fitting.

For implementation, the main difference of our \textit{TTD} training algorithm with traditional training is we add a dynamic targeted dropout layer after each convolutional layer. 
	During the forward phase, the targeted dropout layer will dropout the non-important feature map components (channels and columns).
	This can be done by conducting element-wise multiplication with the generated attention binary mask in Eq.~\ref{eq:mask}:
	\begin{equation}
	\small
		\begin{split}
			& F' = F \otimes M_{spatial}(h, w), \\
			& F'' = F' \otimes M_{channel}(c),
		\end{split}
		\label{eq:spatial_attention}
	\end{equation}
	\normalsize
	where $\otimes$ denotes the element-wise multiplication.
	During multiplication, the mask values will be broadcasted: spatial attention coefficients will be broadcasted and multiply with every entry along the channel dimension, and vice versa.
	$F''$ is the final feature map output with targeted channel-wise and column-wise sparsity.
	During the backward phase, the dynamic dropout layer will just conduct the regular back-propagation without any specific operations.

By introducing the attention-based targeted dropout effect, the \textit{TTD} training will gradually relieve the model's prediction dependency on less-important features (since they are often dropped during training), but increase their focus on most important ones. 
	Therefore, the dynamic pruning of those non-important feature components will induce minimum or no effects to the model accuracy during test-phase inference.


\begin{table*}[]
\renewcommand\arraystretch{1.0}
\centering
\caption{Experiment Results on CIFAR and ImageNet Datasets.}
\setlength{\tabcolsep}{3.7mm}{
\begin{tabular}{cccccccc}
\toprule[1pt]
\begin{tabular}[c]{@{}c@{}}CNN \\ Models\end{tabular} & \begin{tabular}[c]{@{}c@{}}Pruning \\ Methods\end{tabular} & \begin{tabular}[c]{@{}c@{}}Baseline \\ Accuracy(\%)\end{tabular} & \begin{tabular}[c]{@{}c@{}}Baseline \\ FLOPs\end{tabular} & \begin{tabular}[c]{@{}c@{}}Final \\ FLOPs\end{tabular} & \begin{tabular}[c]{@{}c@{}}FLOPs\\ Reduction(\%)\end{tabular} & \begin{tabular}[c]{@{}c@{}}Final\\ Accuracy(\%)\end{tabular} & \begin{tabular}[c]{@{}c@{}}Accuracy\\ Drop(\%)\end{tabular} \\ \midrule[1pt]
\multirow{5}{*}{\begin{tabular}[c]{@{}c@{}}VGG16\\ (CIFAR10)\end{tabular}} 
 & L1 Pruning*~\cite{filter_prune} & 93.3 & - & 2.06E+08 & 34.2 & 93.4 & \textbf{-0.1} \\ \cline{2-8} 
 & Taylor Pruning*~\cite{taylor} & 93.3 & - & 1.85E+08 & 44.1 & 92.3 & 1.0 \\ \cline{2-8} 
 & GM Pruning*~\cite{gm} & 93.6 & - & 2.11E+08 & 35.9 & 93.2 & 0.4 \\ \cline{2-8} 
 & FO Pruning*~\cite{fo} & 93.4 & - & 1.85E+08 & 44.1 & \textbf{93.3} & 0.1 \\ \cline{2-8} 
 & \textbf{Proposed} & 93.3 & 3.13E+08 & \textbf{1.46E+08} & \textbf{53.5} & 93.1 & 0.2 \\ \midrule[1pt]
\multirow{4}{*}{\begin{tabular}[c]{@{}c@{}}ResNet56\\ (CIFAR10)\end{tabular}} 
 & L1 Pruning*~\cite{filter_prune} & 93.0 & - & 0.91E+08 & 27.6 & 93.1 & -0.1 \\ \cline{2-8} 
 & Taylor Pruning*~\cite{taylor} & 92.9 & - & 0.71E+08 & 43.0 & 92.0 & 0.9 \\ \cline{2-8} 
 & FO Pruning*~\cite{fo} & 92.9 & - & \textbf{0.71E+08} & \textbf{43.0} & \textbf{93.3} & \textbf{-0.4} \\ \cline{2-8} 
 & \textbf{Proposed} & 93.0 & 1.28E+08 & 0.80E+08 & 37.4 & 93.2 & -0.2 \\ \midrule[1pt]
\multirow{5}{*}{\begin{tabular}[c]{@{}c@{}}VGG16\\ (CIFAR100)\end{tabular}} 
 & L1 Pruning*~\cite{filter_prune} & 73.1 & - & 1.96E+08 & 37.3 & 72.3 & 0.8 \\ \cline{2-8} 
 & Taylor Pruning*~\cite{taylor} & 73.1 & - & 1.96E+08 & 37.3 & 72.5 & 0.6 \\ \cline{2-8} 
 & FO Pruning*~\cite{fo} & 73.1 & - & 1.96E+08 & 37.3 & 73.2 & {-0.1} \\ \cline{2-8} 
 & \textbf{Proposed: Setting-1} & 73.1 & 3.13E+08 & \textbf{1.87E+08} & \textbf{40.4} & \textbf{73.2} & \textbf{-0.1} \\  \cline{2-8}
 & \textbf{Proposed: Setting-2} & 73.1 & 3.13E+08 & \textbf{1.72E+08} & \textbf{44.9} & 72.9 & 0.2 \\\bottomrule[1pt]
\multirow{5}{*}{\begin{tabular}[c]{@{}c@{}}VGG16\\ (ImageNet100)\end{tabular}} 
 & L1 Pruning*~\cite{filter_prune} & 78.5 & - & 0.76E+10 & 50.6 & 76.6 & 0.8 \\ \cline{2-8} 
 & Taylor Pruning*~\cite{taylor} & 78.5 & - & 0.76E+10 & 50.6 & 77.3 & 0.6 \\ \cline{2-8} 
 & FO Pruning*~\cite{fo} & 78.5 & - & 0.76E+10 & 50.6 & {79.5} & {-1.0} \\ \cline{2-8} 
 & \textbf{Proposed: Setting-1} & 78.5 & 1.52E+10 & \textbf{0.74E+10} & \textbf{51.2} & \textbf{79.6} & \textbf{-1.1} \\  \cline{2-8}
 & \textbf{Proposed: Setting-2} & 78.5 & 1.52E+10 & \textbf{0.69E+10} & \textbf{54.5} & 79.4 & -0.9 \\\bottomrule[1pt]
\end{tabular}
\leftline{* indicates the methods' performance is refered from~\cite{fo, gm}.}
}
\label{table:cifar}
\vspace{-7mm}
\end{table*}

\subsection{Layer Sensitivity Analysis and Dropout Ratio Ascent}
\label{sec:layer_sensitivity}

\begin{figure}[!tb]
	\centering
	\includegraphics[width=3.3in]{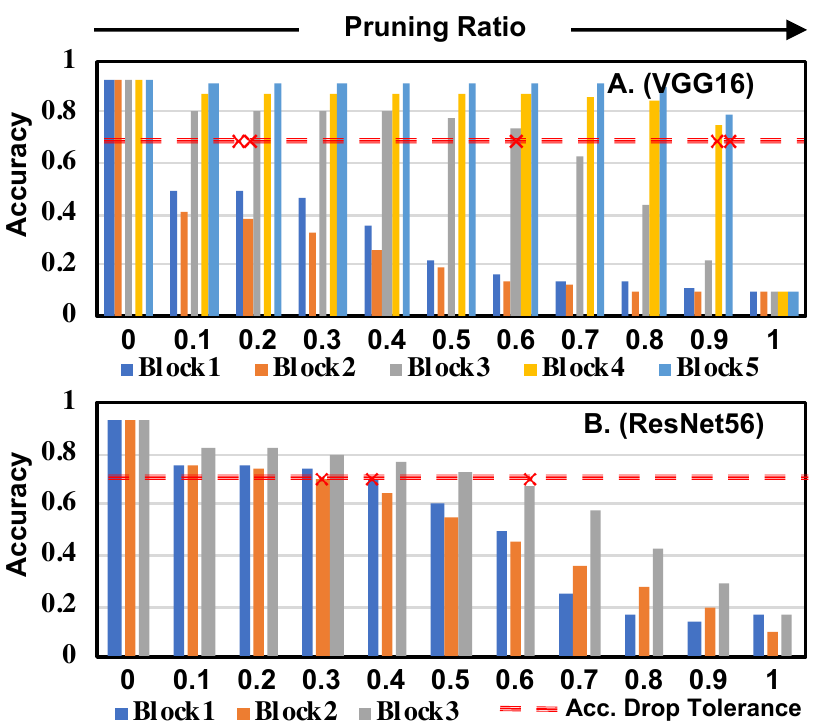}
	\vspace{-3mm}
	\caption{Block Sensitivity Analysis: Accuracy drop trends are different for different blocks, thus they should be set with different pruning ratios.}
	\label{fig:block_test}
	\vspace{-5mm}
\end{figure}

The \textit{TTD} algorithm introduces the targeted dropout into the training process to enable the dynamic pruning during test. 
	Whereas, different layers of a model can have varied amount of redundancy. This requires the targeted dropout ratios to be carefully tuned otherwise it will greatly hurt the model convergence speed and final accuracy. 
	Therefore, we draw some experience from previous static pruning works and follow the layer sensitivity analysis practice to set the dropout ratio for different layers. 

Take VGG16 network as an example. VGG16 has 5 convolutional blocks with [2, 2, 3, 3, 3] convolutional layers and one 2x2 MaxPooling layer at the end of each block. 
	To avoid massive hyper-parameter tuning and to maintain policy consistency with block-wise ResNet structure, we analyze the average block sensitivity and set an aggressive dropout upper bound for each block.
	For example, Fig.~\ref{fig:block_test} shows the block sensitivity analysis for VGG and ResNet. 
	Take VGG as an example: An aggressive pruning ratio per block e.g. [0.2, 0.2, 0.6, 0.6, 0.9] can cause the pruned model's accuracy dropping to less than 70\%, which can be hardly recovered back.
	Therefore, we set this threshold as the upper bound pruning ratio, and then use dropout ratio ascent during \textit{TTD} training.
	The dropout ratio will start with a warm-up ratio, for example 0.1 for each block. 
	After the model converges during the current ratio, we will ascent the ratio for each block with a small step-size (e.g. 0.05) to try reaching the maximum pruning ratio. 
	And the training will stop when the target pruning ratio and a satisfying accuracy is achieved.

After $TTD$ training, the model is then fully-prepared for dynamic pruning with the same ratio during test inference. 
	Therefore, our method doesn't require any further fine-tuning or retraining process, which is another great advantage.









\section{\bf{Experimental Evaluation}}
\label{sec:exp}
In this section, we evaluate our proposed dynamic pruning methods on various of CNN models (e.g. VGG16, and ResNet) on CIFAR10/100, as well as ImageNet. 

\subsection{Experimental Setup}

Our experiments are conducted using deep learning framework PyTorch. 
	On CIFAR10/CIFAR100 dataset, we use the similar data augmentation including random horizontal flip, random crop and 4-pixel padding.
	By default, we use the cosine learning rate decaying~\cite{cosine} (0.1$\rightarrow$0) for $TTD$ training process of all models.
	Several state-of-the-art static pruning methods are chosen as baseline for comparison, including $\ell_1$-norm Pruning~\cite{filter_prune}, Taylor Pruning~\cite{taylor}, Geometric Mean Pruning (GM)~\cite{gm}, Function-Oriented Pruning (FO)~\cite{fo}.

\subsection{Experimental Results on CIFAR and ImageNet}


\paragraph{VGG16 on CIFAR10}

The first model we evaluated is the VGG16 on CIFAR10.
	VGG16 has 13 convolutional layers in 5 blocks.
	For each block, there are 2-2-3-3-3 layers with 64-128-256-512-512 filters (of 3x3 filter size) per layer.
	The experimental results are shown in Table~\ref{table:cifar}. 
	On this model, the best channel pruning ratio per block we find is [0.2, 0.2, 0.6, 0.9, 0.9], which matches the sensitivity analysis in Sec.~\ref{sec:layer_sensitivity}. 
	Beyond this ratio, the accuracy can hardly be compensated by TTD training.
	Since the feature map spatial size is too small (e.g. the last 9 layers' feature map size are ranged from 8x8 to 2x2), pruning spatial columns on VGG16 always brings unrecoverable accuracy drop. 
	Therefore, spatial pruning ratio for this model is set to 0 for all layers. 

Comparing our pruning ratio with previous work, our channel pruning ratios [0.2, 0.2, 0.6, 0.9, 0.9] greatly outnumber the previous state-of-the-art static methods: The best FO pruning~\cite{fo} can only achieve [0.17, 0.1, 0.1, 0.45, 0.65]. 
	This proves our hypothesis that our dynamic method can aggressively and accurately remove more dynamic per-input redundancy which static method fails to.
As a result, our best performance of dynamic pruning is 53.5\% FLOPs reduction with 0.2\% accuracy drop, 9.4\% higher FLOPs reduction than the best FO Pruning with only 0.1\% accuracy difference.

\paragraph{ResNet56 on CIFAR10}

The next model we evaluated is ResNet56 on CIFAR10. 
	Different from VGG16 with wide layers, ResNet56 has three groups containing 16 convolutional layers per group. And it has at most 64 convolutional filters in one layer.
	Thus, the channel redundancy in ResNet56 is relatively limited compared to VGG16.
	By contrast, its feature map size are from 32x32 to minimum 8x8, which can contain more redundancy. 
	Therefore, for this network, we prune less channels but more spatial columns. 
	Due to the skip connection, we need to maintain the same size of input and output channels of even layers in every group. 
	Therefore, the dynamic pruning is only conducted in the odd layers in the group.
	The dynamic pruning setting for this network is channel-wise pruning ratio: [0.3, 0.3, 0.6], and spatial-wise pruning ratio: [0.6, 0.6, 0.6].
	The final performance is 37.4\% FLOPs reduction with slight 0.2\% accuracy improvement, which is comparable with previous FO pruning results. 

\paragraph{VGG16 on CIFAR100}

We also test our dynamic pruning method's performance using VGG16 on CIFAR100.
	One conservative pruning setting (Setting-1) is [0.2, 0.2, 0.2, 0.8, 0.9] channel-wise ratio, with zero spatial pruning ratio for all layers due to the same reason of small feature map size.
	Compared to all baseline methods, this setting could achieve the highest FLOPs reduction (40.4\%) as well as the highest accuracy (73.2\%). 
	We also conduct a more aggressive pruning setting (Setting-2) with [0.3, 0.2, 0.2, 0.9, 0.9] channel-wise ratio and zero spatial ratio. 
	In this setting, a 0.5\% accuracy drop is observed but we could push the FLOPs reduction to 44.9\%, which is 4.5\% more than Setting-1 model.


\paragraph{VGG16 on ImageNet100}

To validate our method's performance on large-scale image datasets, we test our pruning methods using VGG16 model on ImageNet100 dataset with the same setting as~\cite{fo}. 
	Different from CIFAR datasets, images in ImageNet has a size of 224x224x3. 
	Therefore, the spatial dimension of the feature map in this model is much larger than CIFAR model.
	With the similar FLOPs reduction as the baseline, our setting-1 model brings 51.2\% FLOPs reduction. The setting is [0.1, 0, 0, 0, 0.2] for channel-wise ratio, and [0.5, 0.5, 0.5, 0.5, 0.5] for spatial ratio.
	In Setting-2, we further achieve 54.5\% FLOPs reduction with increased spatial ratio [0.5, 0.5, 0.5, 0.6, 0.6] and without accuracy drop. 
	This further proves our method's good capability in removing model redundancy on large-scale datasets.

\subsection{The Redundancy Existence in Various Dimension}

\begin{figure}[!b]
	\centering
	\vspace{-5mm}
	\includegraphics[width=3.3in]{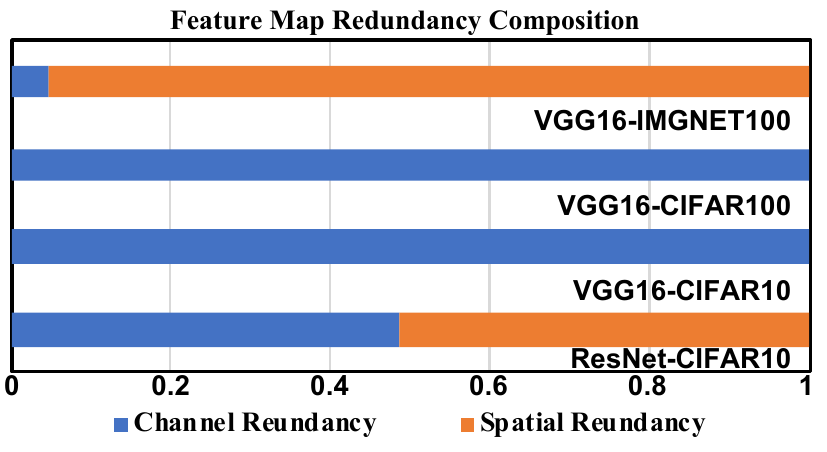}
	\vspace{-3.5mm}
	\caption{The redundancy composition varies in channel and spatial dimensions.}
	\label{fig:dimension}
\end{figure}

In above experiments, we calculate the FLOPs reduction all together. 
	But if we calculate the channel and spatial redundancy separately, we could find that the model redundancy can exist in very arbitrary dimensions.
	As shown in Fig.~\ref{fig:dimension}, on ImageNet VGG16, the removed channel-wise redundancy only accounts for 2.4\% FLOPs reduction but the spatial-column redundancy accounts for 52.1\%.
	This implies that about half of feature map redundancy exists in the spatial dimension.
	However, remind that with the same VGG structure on CIFAR with 32x32 image size, the major redundancy is removed in a opposite channel-wise manner (all spatial ratio is zero).
	While for ResNet56 model, a moderate amount of channel redundancy (18.2\%) as well as similar amount of spatial redundancy (19.2\%) can be removed simultaneously.

Such difference highlights the fact that the feature map redundancy can exist in various of dimensions with different input scales and model structures.
	Therefore, the previous channel-only pruning can hardly remove the spatial redundancy in the feature maps. 
	By comparison, our new designed spatial column pruning and its combination with channel pruning method can conduct much more flexible and thorough feature map redundancy removal, which thus achieve the best performance in most test settings.

\section{\bf{Conclusion}}

In this work, we propose a dynamic feature map pruning method based on attention mechanism. 
	Specifically, we consider both channel and spatial column redundancies and removed them by our proposed dynamic pruning methods.
	Meanwhile, to enhance the pruning performance, we propose a training with targeted dropout method to further improve the pruned model accuracy. 
	Extensive experiments validate our method's effectiveness and show that our method achieves better performance than previous state-of-the-art methods.

\end{document}